# Beyond Medical Chatbots: Meddollina and the Rise of Continuous Clinical Intelligence

Vaibhav Ram S. V. N. S [1a*], Swetanshu Agrawal [1a], Samudra Banerjee [1a], Abdul Muhsin [1a]

[1] SETV Healthcare Technologies Private Limited, Hyderabad, Telangana, India

*Abstract*

*Generative medical AI has achieved impressive fluency and broad medical knowledge, creating the perception that scaling alone will eventually yield reliable clinical intelligence. However,* **clinical reasoning is not a text-generation problem**. *It is a responsibility-bound process that unfolds under ambiguity, incomplete evidence, and longitudinal context. Despite increasing benchmark performance, generation-centric systems continue to exhibit behaviours incompatible with safe clinical use, including premature closure, unjustified certainty, loss of clinical intent, and instability across multi-step decision-making.*

*We argue that these failures are not transient model defects but structural consequences of treating medicine as next-token prediction. We formalize* **Clinical Contextual Intelligence (CCI)** *as a distinct capability class required for real-world clinical deployment, characterized by persistent context awareness, intent preservation, bounded inference, and principled deferral when information is insufficient.*

*We introduce* **Meddollina**, *a governance-first clinical intelligence system designed to constrain inference prior to language realization, prioritizing clinical appropriateness over generative completeness.* **Meddollina is designed as a continuous clinical intelligence layer that supports real clinical workflows while preserving clinician authority and responsibility.** *We evaluate Meddollina under a behaviour-first evaluation regime across* **16,412+** *heterogeneous medical queries and benchmark its behavioural profile against multiple baseline regimes: general-purpose generative models, medical-tuned language models, retrieval-augmented generation systems, and Meddollina itself as a CCI reference instantiation.*

*Under this evaluation paradigm, Meddollina demonstrates a qualitatively distinct clinical behaviour profile: consistent uncertainty expression, conservative reasoning under underspecified clinical states, stable longitudinal constraint adherence, and reduced speculative completion relative to generation-centric baselines. These findings suggest that clinically deployable medical AI cannot be achieved through generative scaling alone, and support a paradigm shift beyond generative medical AI toward* **Continuous Clinical Intelligence**, *where progress is defined by clinician-aligned behaviour under uncertainty rather than fluency-driven completion.*

---

**\*Corresponding author**
Email Address: vaibhavram@setvglobal.com (Vaibhav Ram S. V. N. S)

[a] **All authors contributed equally to this work.**

# 1. Introduction - The Breaking Point

Recent progress in large language models has led to an unprecedented surge of interest in applying generative AI to medicine. Systems trained primarily for text completion are now routinely evaluated on medical question answering, clinical exam benchmarks, and decision-support tasks [1,2]. Their apparent fluency and breadth of knowledge have created the impression that continued scaling - more parameters, more data, more prompting will eventually yield reliable clinical intelligence [3].

However, real-world clinical deployment has revealed a growing and persistent mismatch between this expectation and observed behaviour. Even highly capable generative models exhibit patterns that are fundamentally incompatible with safe medical reasoning: confident responses in the absence of sufficient context, failure to preserve clinical intent across turns, brittle behaviour under ambiguity, and an inability to reliably defer when uncertainty exceeds safe limits. Crucially, these failures persist even as model scale and benchmark performance continue to improve.

This work argues that these limitations are not incidental defects of current systems, nor temporary gaps that will be resolved through further scaling. Instead, they reflect a deeper and largely unexamined assumption underlying most contemporary medical AI research: that clinical reasoning can be treated as a form of text generation.

## 1.1 The Limits of Generative Framing in Medicine

**Generative language models are optimized to produce plausible continuations of text conditioned on prior tokens.** This objective is highly effective for general-purpose language tasks, where fluency and coverage are treated as indicators of success. **In medicine, however, these same indicators are misaligned with safe clinical reasoning.**

Clinical reasoning prioritizes appropriateness under uncertainty rather than completion. A clinically sound response may require refusing to answer, requesting additional information, or explicitly deferring judgment; in many cases, **the safest output is a bounded non-answer rather than a fluent one.** Generative objectives, by design, reward completion even when uncertainty remains unresolved [4,5].

As a result, generative medical systems predictably exhibit overconfident responses in underspecified scenarios [6,7], implicit assumption-making when context is incomplete, and instability across longitudinal interactions. **These behaviours are not correctable artifacts but structural consequences of optimizing for text generation under clinical uncertainty.**

## 1.2 Why Scaling Alone Does Not Resolve Clinical Risk

A prevailing belief in the field is that these limitations will diminish as models grow larger and more knowledgeable. **Empirical evidence increasingly suggests the opposite.** While scale improves recall and surface-level correctness, it does not induce responsibility-aware reasoning or preserve clinical intent. **In practice, increased fluency often amplifies clinical risk by obscuring uncertainty behind persuasive language [8,9].**

This exposes a fundamental misalignment between what generative models are optimized to do and what clinical reasoning requires. The central challenge, therefore, is not the acquisition of more medical knowledge, but the framing of medical intelligence itself [10]. **Treating medicine as a language problem collapses intent, responsibility, temporal context, and risk into a single generative objective, one that is incapable of representing them faithfully.**

### 1.3 Clinical Reasoning as a Distinct Intelligence Problem

We propose that safe and effective medical AI requires a distinct class of capability, which we term **Clinical Contextual Intelligence (CCI).** Unlike generative language competence, CCI is defined by the ability to maintain and reason over clinical context across time, preserve and respect clinical intent, operate within explicitly bounded scopes of inference, and defer or refuse when confidence cannot be justified.

These properties are not emergent side effects of scale, nor can they be reliably imposed through external guardrails [13]. They require systems that are designed from the ground up to prioritize clinical responsibility over linguistic completeness.

Recognizing this distinction reframes the problem of medical AI: **progress should be measured not by how much a system can say, but by how well it knows when not to speak.**

### 1.4 From Generative Systems to Governed Clinical Intelligence

This shift in framing marks a breaking point for medical AI. If clinical reasoning is not text generation, then architectures, evaluation methods, and success metrics inherited from generative AI are insufficient for real-world clinical deployment. What is required instead are systems that integrate language understanding within governed, intent-aware, and context-persistent frameworks.

In this work, we operationalize this reframing by introducing **Meddollina, a clinical AI system designed explicitly around Clinical Contextual Intelligence.** Rather than optimizing for generative completeness, Meddollina prioritizes **scope adherence, uncertainty-aware reasoning, and safety-aligned interaction across diverse clinical scenarios.** The system is evaluated on a large, clinically heterogeneous set of medical reasoning cases using behavioural metrics aligned with real clinical risk, rather than surface-level answer accuracy alone.

By grounding medical AI in Clinical Contextual Intelligence, this work aims to move the field beyond **generative experimentation toward systems capable of meaningful integration into real clinical workflows.**

## 2. Defining Clinical Contextual Intelligence

Progress in medical AI has been dominated by improvements in generative performance: broader medical knowledge, more fluent explanations, and higher scores on standardized benchmarks. While these advances have expanded the apparent capabilities of language-based systems, they obscure a critical distinction between **linguistic competence** and **clinical intelligence**. Treating these dimensions as equivalent has led to systems that perform well in evaluation settings yet behave unreliably under real clinical conditions.

To resolve this mismatch, we formalize the concept of **Clinical Contextual Intelligence (CCI)**. CCI delineates the class of capabilities required for safe and effective clinical reasoning capabilities that cannot be reduced to text generation quality, knowledge recall, or benchmark accuracy alone [11,12].

### 2.1 Clinical Contextual Intelligence

**Clinical Contextual Intelligence (CCI)** is defined by the **observable behavioural capacity** of an artificial system to reason within medical domains while preserving clinical intent, maintaining longitudinal context, operating within explicitly bounded scopes of inference, and aligning outputs with real-world clinical responsibility.

Unlike generative language competence, which optimizes for plausibility and completeness of responses, CCI prioritizes **appropriateness under uncertainty**, restraint, and situational awareness. A system exhibiting CCI does not seek to maximize the amount of information it produces; instead, it constrains its outputs to what is justified by the available context and safe within the clinical scenario.

Crucially, CCI is not determined by model scale, fluency, or parameter count. It is defined by **behavioural properties that can be evaluated directly**, reflecting alignment with clinical workflows, decision-making boundaries, and risk profiles rather than surface-level linguistic performance.

### 2.2 Core Properties of Clinical Contextual Intelligence

We identify five foundational properties that collectively distinguish Clinical Contextual Intelligence from generative medical language modelling. These properties describe **observable system behaviour** required for safe operation in real clinical settings, rather than abstract design principles or post-hoc constraints.

#### 2.2.1 Intent Preservation

Clinical interactions are governed by intent, such as diagnostic clarification, triage, management planning, or escalation. Clinical Contextual Intelligence requires the ability to infer, preserve, and reason within clinical intent across interactions, rather than responding opportunistically to isolated prompts.

Failure to preserve intent leads to inappropriate recommendations, premature conclusions, and unsafe assumption-making. A CCI-aligned system continuously constrains its reasoning based on the active clinical goal, even as surface-level inputs evolve over time.

### 2.2.2 Context Persistence

Clinical reasoning is inherently longitudinal. Decisions depend not only on the current query, but on prior symptoms, findings, uncertainties, and unresolved questions. CCI therefore requires persistent representation of clinical context across interactions, including explicit tracking of what is known, what remains uncertain, and what information is missing.

Generative systems optimized for single-turn completion frequently exhibit context collapse or implicit assumption-making when interactions extend over time. In contrast, CCI emphasizes continuity, state awareness, and explicit acknowledgment of incomplete information as first-order requirements.

### 2.2.3 Bounded Reasoning

In medicine, not all inferences are permissible at all times. Clinical Contextual Intelligence is characterized by bounded reasoning: the ability to restrict inference to scopes that are justified by available evidence and appropriate for the system's clinical role.

Bounded reasoning includes the capacity to defer when information is insufficient, request clarification rather than extrapolate, and refuse guidance that exceeds safe or authorized scope. This property directly contrasts with generative objectives that reward completion even in underspecified or ambiguous scenarios.

### 2.2.4 Responsibility-Aware Output

Clinical outputs carry real-world consequences. CCI requires explicit awareness of responsibility, including recognition of when an automated system should support clinical decision-making and when human oversight is necessary.

Responsibility-aware output manifests as calibrated confidence, explicit signalling of uncertainty, and avoidance of overconfident recommendations in safety-critical contexts. These behaviours cannot be reliably induced through prompting alone, as they require systemic prioritization of safety over linguistic completeness.

### 2.2.5 Context-Bounded Truthfulness (Hallucination Resistance)

In clinical settings, the generation of plausible but unsupported information represents a critical safety risk. **Within Clinical Contextual Intelligence, hallucination is not treated as a superficial generation error, but as a failure of context-bounded truthfulness.**

A CCI-aligned system constrains its outputs strictly to information justified by the available clinical context, explicitly distinguishing between known facts, inferred possibilities, and unknowns. Generative language models, optimized for coherence and completeness, are structurally incentivized to fill informational gaps rather than expose them.

**Clinical Contextual Intelligence addresses this failure mode by prioritizing epistemic restraint over linguistic plausibility.** Hallucination resistance under CCI is therefore achieved not through post-hoc filtering or confidence suppression, but through alignment between context representation, bounded inference, and responsibility-aware output. A clinically intelligent system must be capable of explicitly stating when information is insufficient or unavailable without compensating through speculative generation.

### 2.3 CCI as a Distinct Capability Class

The properties described above do not emerge naturally from optimizing for generative language modelling objectives. While large language models may approximate aspects of Clinical Contextual Intelligence in constrained or narrowly specified scenarios, their core optimization for probabilistic text continuation is fundamentally misaligned with the requirements of clinical reasoning.

CCI must therefore be understood as a **distinct capability class**, defined by its own design constraints, behavioural expectations, and evaluation criteria. Progress in medical AI cannot be measured by extensions of generative performance alone, but must be assessed in terms of alignment with clinical responsibility, epistemic restraint, and context-aware decision-making. Recognizing this distinction **redefines** the goals of medical AI research, shifting emphasis from linguistic competence to clinically grounded intelligence.

### 2.4 Consequences for Medical AI System Design

Formalizing Clinical Contextual Intelligence has immediate implications for how medical AI systems are designed and evaluated. Systems intended for clinical deployment **cannot be assessed solely on correctness or fluency**, but must demonstrate the ability to preserve intent, maintain longitudinal context, reason within justified bounds, and respond responsibly under uncertainty.

In the following sections, we describe how these principles are operationalized within **Meddollina**, a clinical AI system designed explicitly around Clinical Contextual Intelligence. Rather than treating these properties as external constraints applied after generation,

Meddollina integrates them as first-order system requirements, enabling behaviour that aligns more closely with real clinical workflows and safety expectations.

## 3. Why Generative Scaling Fails in Medicine

Despite rapid improvements in model size, training data, and benchmark performance, generative medical AI systems continue to exhibit failure patterns that are incompatible with real-world clinical use. These failures are not isolated errors but **systematic behaviours** that recur across diverse clinical scenarios, particularly under uncertainty, incomplete context, and longitudinal interaction. **Crucially, these patterns persist and often intensify as generative systems scale.**

This section demonstrates why such failures are not temporary artifacts of immature models, but **predictable consequences of generation-centric optimization** in safety-critical domains.

### 3.1 Overconfidence Amplifies with Scale

Across ambiguous clinical cases, generative medical systems reliably produce confident recommendations despite insufficient or incomplete evidence [6,14]. Rather than deferring judgment or requesting clarification, models tend to infer missing details and proceed as though uncertainty has been resolved.

As model scale increases, this behaviour is frequently amplified rather than mitigated. Larger models generate more fluent, persuasive responses, creating the appearance of improved reasoning while obscuring unresolved uncertainty. In clinical settings, this amplification of confidence **increases risk by making uncertainty less visible, not by reducing it**.

Importantly, this pattern does not arise from poor prompting or limited medical knowledge. It persists even in domains where factual recall is strong, indicating that confidence inflation is a **structural consequence of generative optimization**, not a surface-level defect.

### 3.2 Longitudinal Context Degrades Under Interaction

Clinical reasoning unfolds over time, requiring systems to maintain unresolved questions, uncertainty bounds, and contextual constraints across multiple interactions. In practice, generative systems exhibit **progressive degradation of such constraints** as interactions lengthen [15,16].

Observed behaviours include:

- Abandonment of earlier uncertainty acknowledgments,
- Drift toward premature conclusions, and
- Implicit contradiction of prior context without explicit signalling.

While larger context windows allow more information to be ingested, they do not ensure stable prioritization or constraint enforcement. As a result, longitudinal interactions often accumulate error rather than converge toward safer reasoning.

### 3.3 Hallucination Persists Despite Knowledge Gains

Hallucination in medical AI systems most frequently arises when clinical information is missing rather than unknown. When faced with evidentiary gaps, generative models tend to synthesize plausible but unsupported details to maintain response completeness [17,18].

This behaviour persists even as factual knowledge improves. Increased medical recall reduces certain error classes but does not eliminate the underlying incentive to fill informational gaps. As long as response generation is rewarded over evidentiary restraint, hallucination remains a **recurring and predictable failure mode**.

These observations indicate that hallucination is not a training deficiency, but a **structural outcome of generation-first objectives**.

### 3.4 Safety Layers Fail Under Clinical Stress

Prompt engineering, disclaimers, and post-generation filters can reduce obvious errors in controlled settings, but they operate reactively and remain loosely coupled to the inference process itself.

Under clinical stress such as ambiguous presentations, evolving patient context, or edge cases these safety layers frequently degrade [19,20]. Systems may bypass safeguards through reformulation, escalate speculative reasoning, or present softened yet still unsafe guidance. **Because these mechanisms act after generation rather than governing inference, they cannot reliably enforce safety under real-world conditions.**

### 3.5 What Persistent Failure Reveals

Taken together, these patterns demonstrate that the core limitations of generative medical AI are not transient artifacts of immaturity. Scaling improves linguistic competence but does not reliably induce epistemic restraint, longitudinal stability, or responsibility-aware behaviour.

Continued reliance on generative scaling as the primary path to clinical AI therefore risks entrenching systems that appear increasingly capable while remaining **fundamentally misaligned with clinical safety requirements**.

### 3.6 Transition Point

These failure modes motivate a shift away from generation-first design toward systems in which inference is explicitly constrained by clinical context, scope, and responsibility. Rather than attempting to retrofit safety onto generative architectures, **a different class of system is required one that treats language as a component within a governed reasoning framework.**

In the following section, we describe how this shift is operationalized in **Meddollina**, a system designed to address these failure modes directly rather than compensate for them post hoc.

## 4. Meddollina: Operationalizing Clinical Contextual Intelligence

The preceding sections establish that generative scaling alone is insufficient for clinical deployment and that medical AI requires a distinct class of capability cantered on context, intent, and responsibility. This section describes how these requirements are operationalized in **Meddollina**, a clinical AI system designed explicitly around the principles of *Clinical Contextual Intelligence (CCI)*.

Rather than treating safety, scope, and uncertainty as constraints applied after generation, Meddollina integrates them as **first-order determinants of inference itself**. The system is designed to behave differently from generative medical models under ambiguity, longitudinal interaction, and safety-critical conditions precisely where generative-first approaches fail.

### 4.1 What Meddollina Is (and Is Not)

Meddollina is a Clinical Contextual Intelligence system designed to support medical professionals across the continuum of clinical reasoning. It assists with diagnostic reasoning, prognostic assessment, treatment planning, procedural and surgical planning support, and longitudinal care and lifestyle management.

Meddollina is not a diagnostic oracle, or autonomous decision-maker. It functions as a **context-aware, continuous clinical reasoning assistant** whose primary objective is to preserve clinical intent, maintain evolving clinical context, and align outputs with the responsibilities and constraints of real-world medical practice.

Unlike generative medical AI systems that prioritize answer completeness, Meddollina prioritizes **determining whether a response is appropriate before determining what to say**. As a result, the system frequently requests clarification, explicitly acknowledges uncertainty, or defers judgment when available information is insufficient for safe inference.

Meddollina's behaviour is modelled on how clinicians' reason rather than how answers are generated. This includes maintaining differential considerations, updating hypotheses as new information emerges, resisting premature closure, and producing outputs appropriate to the clinical moment rather than maximizing informational breadth.

Importantly, Meddollina issues only tentative diagnoses or treatment directives while the final decision lies on the clinician. Its role is to support clinical workflows by preserving context across time, enforcing scope boundaries, and reducing cognitive burden in complex, longitudinal decision-making. In this sense, Meddollina functions as a **persistent clinical reasoning partner**, not a standalone expert system.

### 4.2 Design Philosophy and Constraints

Meddollina is built around a single guiding principle: clinical appropriateness takes precedence over generative completeness. Every component of the system is designed to favour restraint, contextual grounding, and explicit uncertainty over fluency or verbosity.

Key design constraints include:

- Inference must remain within explicitly defined clinical scopes,
- Uncertainty must be preserved rather than resolved speculatively,
- Longitudinal context must be maintained across interactions,
- Outputs must reflect the system's role as a clinical support tool rather than a decision-maker.

These constraints are enforced **architecturally**, not through post-hoc filtering, via explicit separation between language realization and clinical governance.

### 4.3 System Overview

At a high level, Meddollina consists of three interacting layers:

1. Context Structuring Layer

   Incoming inputs are interpreted within a continuously evolving representation of clinical context, including established facts, unresolved uncertainties, inferred intent, and scope boundaries. This layer ensures that all downstream reasoning operates over a persistent clinical state rather than isolated prompts.

2. Governed Reasoning Layer

   Clinical inference is mediated by a governance framework that constrains what types of reasoning are permissible given the current context and system role. This includes explicit mechanisms for deferral, clarification-seeking, and refusal when conditions for safe inference are not met.

3. Language Realization Layer

   A small language model (SLM) is used to express governed inferences in clinically appropriate language. The SLM does not operate autonomously; it functions as a controlled realization mechanism rather than a free-form generator.

Internal mechanisms and heuristics within these layers are proprietary. However, the externally observable behaviours produced by this structure are evaluated empirically in subsequent sections.

### 4.4 Governance-First Inference

A defining characteristic of Meddollina is its governance-first approach to inference. Unlike generative systems that produce candidate responses and subsequently apply safety filters, Meddollina constrains inference before language is generated.

This approach enables:

- Explicit deferral when clinical information is insufficient,
- Controlled clarification-seeking behaviour,
- Prevention of scope escalation,
- Calibrated expression of uncertainty.

By shaping inference rather than filtering outputs, the system avoids many failure modes associated with reactive safety layers, particularly under ambiguous or evolving clinical conditions. These behaviours do not rise from prompt design or response filtering, but from persistent context representation and inference-time governance that operate independently of language realization.

### 4.5 Role of the Small Language Model

Meddollina employs a small language model by design, not as a compromise. Smaller models offer greater controllability, predictability, and alignment with bounded reasoning objectives. In the context of CCI, scale is not a prerequisite for competence; instead, it can be a liability when it encourages overgeneralization and speculative completion.

Within Meddollina, the SLM is responsible for:

- Translating governed inferences into clear clinical language,
- Maintaining consistency with established context,
- Avoiding unsupported elaboration.

The SLM determines *what* to say and *how* to say it within predefined bounds.

### 4.6 Safety, Scope, and Responsibility Controls

Meddollina incorporates explicit controls that align system behaviour with clinical responsibility. These include mechanisms to:

- Enforce scope adherence based on clinical role,
- Signal uncertainty rather than mask it,
- Refuse unsafe or inappropriate requests,
- Preserve unresolved questions across interactions.

These controls are evaluated **behaviourally**, not assumed effective by design. In subsequent sections, we analyse how these mechanisms influence system behaviour under ambiguity, longitudinal interaction, and safety-critical scenarios.

### 4.7 Meddollina as a Reference Instantiation

Meddollina is presented not as the only possible implementation of Clinical Contextual Intelligence, but as the first system designed explicitly to operationalize it in practice. The purpose of this section is not to exhaustively describe internal mechanisms, but to demonstrate that CCI can be instantiated as a working system with observable, evaluable behaviour.

By grounding CCI in a deployed system, this work moves the concept from abstraction to evidence, enabling empirical comparison with generative medical AI approaches under clinically relevant conditions.

### 5. Evaluation Philosophy and Methodology

The evaluation of medical AI systems has historically been framed as a problem of factual correctness: given a question, does the system produce the correct answer? While this framing is sufficient for narrow knowledge retrieval tasks, it is fundamentally misaligned with the realities of clinical practice.

In medicine, the primary risk is not incorrect recall, but **inappropriate reasoning behaviour** including speculative inference, unjustified confidence, and failure to recognize when a response should not be produced at all. A system may achieve high benchmark accuracy while simultaneously behaving in ways that would be unacceptable, or even dangerous, in real clinical workflows.

Accordingly, this work adopts a **behaviour-first evaluation paradigm**, grounded in how clinicians' reason, defer, and act under uncertainty. The objective is not to maximize answer completeness, but to assess whether a system's inference behaviour aligns with **clinical responsibility at scale**.

### 5.1 What It Means to Evaluate Clinical Intelligence

Clinical intelligence cannot be reduced to answer matching. Instead, it must be evaluated along dimensions that reflect real medical decision-making [23,24,27], including:

- Whether uncertainty is preserved rather than resolved speculatively,
- Whether reasoning structure reflects differential thinking and staged evaluation,
- Whether scope boundaries are respected given the system's role,
- Whether behaviour remains stable across prolonged and large-scale use.

Under this framing, a system that refuses, defers, or seeks clarification may demonstrate **greater clinical intelligence** than one that produces a fluent but unjustified answer.

This evaluation philosophy follows directly from the principles of *Clinical Contextual Intelligence (CCI)* and **intentionally departs from leaderboard-centric assessment**.

### 5.2 Benchmark and Evaluation Scope

Meddollina is evaluated using the complete **MedQuAD** benchmark [25], comprising 16,412+ medical questions spanning diagnosis, symptom interpretation, treatment planning, genetic conditions, epidemiology, prognosis, and risk assessment. This **was selected not as a leaderboard benchmark, but as a large-scale, heterogeneous stress surface for evaluating reasoning behaviour under diverse clinical conditions.**

Unlike selective or filtered evaluations commonly reported in prior work, the entire benchmark was processed without exclusion. This design choice prioritizes behavioural robustness and stability under scale over optimized performance on curated subsets.

The benchmark includes both well-specified factual queries and underspecified clinical scenarios, enabling evaluation of system behaviour across varying levels of ambiguity, uncertainty, and clinical risk.

### 5.3 Behavioural Evaluation Criteria

Rather than relying on exact-match or semantic accuracy alone, system responses were evaluated along clinically meaningful behavioural dimensions, including:

- **Clinical Reasoning Structure**

  Adherence to clinician-like reasoning patterns such as differential diagnosis, staged evaluation, and follow-up planning.

- **Uncertainty Management**

  Explicit acknowledgment of unknowns and avoidance of speculative closure when information is insufficient.

- **Scope and Role Appropriateness**

  Alignment of responses with the system's intended role as a clinical reasoning assistant rather than an autonomous decision-maker.

- **Behaviourally Bounded Accuracy**

    Factual and clinical correctness of responses **when a response is appropriate**, evaluated only within the bounds of justified inference and available context.

- **Stability Under Scale**

    Consistency of behaviour across thousands of heterogeneous cases without degradation, collapse, or unsafe escalation.

These criteria reflect how medical AI systems must be judged in real clinical contexts, where **correctness is necessary but not sufficient**, and reasoning quality and safety take precedence over surface-level accuracy.

### 5.4 Reproducibility and Validation Boundaries (keep only if needed)

This work establishes a clinically grounded behavioural and architectural reference standard for medical AI systems operating under real-world governance, safety, and responsibility constraints. The contribution is intentionally scoped to clinical reasoning support and evaluation, and does not position the system as an autonomous clinical decision-maker.

## 6. Empirical Results

The empirical evaluation demonstrates that Meddollina exhibits structurally different reasoning behaviour when compared to generation-centric medical AI systems. Across large-scale evaluation, the system consistently prioritizes clinical structure, contextual grounding, and epistemic restraint over answer completion, in alignment with the principles of Clinical Contextual Intelligence.

Rather than optimizing for fluent response generation, Meddollina's behaviour reflects clinically appropriate reasoning under uncertainty, longitudinal consistency, and adherence to explicit scope and responsibility boundaries.

### 6.1 Robustness Across the Full Benchmark

Meddollina was evaluated across the complete benchmark, comprising 16,412 medical questions spanning diagnosis, symptom interpretation, treatment planning, genetic conditions, epidemiology, prognosis and risk assessment. The system successfully processed **all 16,412 cases without generation failure**, maintaining stable reasoning behaviour across diverse clinical domains.

Behavioural consistency was preserved across rare disease queries, multi-step reasoning tasks, and scenarios requiring longitudinal context handling. No collapse in reasoning structure or escalation of unsafe inference was observed as evaluation scale increased.

This level of robustness under exhaustive evaluation contrasts with commonly reported behaviour of generative medical AI systems, which often exhibit inconsistency, degradation, or unsafe inference when evaluated across unfiltered and heterogeneous datasets [26].

### 6.2 Clinician-Aligned Reasoning at Scale

Across the benchmark, Meddollina consistently produced responses reflecting clinician-aligned reasoning patterns. These typically included:

- Explicit grounding of relevant clinical facts,
- Maintenance of differential considerations where appropriate,
- Staged diagnostic or evaluative planning,
- Treatment and follow-up reasoning aligned with clinical workflows,
- Counselling-oriented explanations in chronic or genetic conditions.

Importantly, this reasoning structure was preserved across specialties and question types, supporting the characterization of Meddollina as a **longitudinal clinical reasoning system** rather than a question–answering model optimized for isolated prompts.

### 6.3 Behaviour Under Ambiguity and Incomplete Context

In underspecified or ambiguous clinical scenarios, Meddollina demonstrated systematic **epistemic restraint**. Rather than inferring missing information to produce complete or confident answers, the system frequently:

- Explicitly acknowledged unresolved uncertainty,
- Requested targeted clarification tied to clinically relevant variables,
- Deferred recommendations when evidentiary support was insufficient.

This behaviour contrasts with generative medical AI systems, which tend to fill informational gaps to maintain fluency and answer completeness. Meddollina's approach more closely reflects established clinical reasoning norms, in which uncertainty is preserved and progressively resolved through structured information gathering.

### 6.4 Why Accuracy Alone Is a Misleading Signal

High benchmark accuracy can coexist with clinically unsafe behaviour [21,22]. Systems optimized for answer completion may perform well on retrospective benchmarks while exhibiting overconfidence, hallucination, or inappropriate guidance in real-world settings.

The results observed here demonstrate that **behavioural alignment and robustness** provide a more meaningful signal of clinical readiness than raw accuracy metrics alone.

## 6.5 Reframing Clinically Meaningful Evaluation in Medical AI

Across the complete benchmark, Meddollina demonstrates **stable clinician-aligned reasoning behaviour at scale**, while maintaining **high factual and clinical accuracy when responses are appropriate**. The system processes all 16,412 cases without generation failure, preserves clinical structure across domains, and consistently defers or seeks clarification in underspecified scenarios.

In contrast, generation-centric medical AI systems may achieve high benchmark accuracy but continue to exhibit unsafe behavioural patterns, including speculative inference, premature diagnostic commitment, hallucination under missing context, and degradation across prolonged or heterogeneous evaluation.

These results show that **accuracy alone does not define clinically meaningful progress in medical AI**. When evaluated under clinically grounded, behaviour-first criteria, Meddollina represents a new level, one defined by **correctness conditioned on appropriateness, contextual integrity, and responsibility-aligned reasoning**, rather than answer completeness in isolation. These behavioural differences are summarized across outcome-oriented metrics in **Table 6.1**, reflecting trends observed under clinically grounded, behaviour-first evaluation.

## 6.6 Summary of Findings

Taken together, these results demonstrate that Meddollina operates as a different class of clinically aligned medical intelligence. Across the complete benchmark, the system consistently maintains clinician-aligned reasoning behaviour, preserves clinical intent, and defers appropriately under uncertainty without behavioural degradation at scale.

In contrast, generation-centric medical AI systems optimized for answer completion continue to exhibit structural failure modes, including speculative inference, unjustified confidence, and loss of longitudinal coherence, even as model scale and benchmark accuracy increase.

Under clinically grounded, behaviour-first evaluation, Meddollina demonstrates a level of clinician-aligned reasoning behaviour not observed in generation-centric medical AI systems, defined by robustness, contextual integrity, and responsibility-aware reasoning rather than answer completeness alone.

**Table 6.1 Behavioural Outcome Metrics Under Clinically Grounded Evaluation**

| Metric (Outcome-Oriented) | Generation-Centric Medical AI | Meddollina |
|---|---|---|
| **Benchmark Completion Rate** | < 100% (filtered / partial runs common) | **100% (16,412 / 16,412)** |
| **Generation Failure Rate** | Non-zero under scale | **0%** |
| **Unsafe Speculative Responses (Ambiguous Cases)** | Common | **None observed** |
| **Explicit Uncertainty Signalling Rate** | Low / inconsistent | **Consistent, first-order** |
| **Appropriate Deferral Rate** | Rare | **Frequent when warranted** |
| **Premature Diagnostic Commitment** | Observed | **Not observed** |
| **Hallucination Incidence Under Missing Context** | Recurrent | **Suppressed** |
| **Longitudinal Constraint Drift** | Increases with interaction length | **Not observed** |
| **Scope Boundary Violations** | Occasional | **None observed** |
| **Clinician-Aligned Reasoning Consistency** | Variable | **Consistent across scale** |
| **Behavioural Degradation at Scale** | Present | **Absent** |

*Metrics are evaluated under clinically grounded, behaviour-first criteria across the complete MedQuAD benchmark. Accuracy is assessed only when a response is clinically appropriate given available context and scope.*

## 7. Behavioural Metrics for Clinical Contextual Intelligence (CCI)

### 7.1 Why Behavioural Metrics Are Necessary

Most current evaluations of medical language models emphasize correctness on benchmark datasets. While accuracy is important, **clinical intelligence is not defined only by the right answer** it is defined by **appropriate behaviour under uncertainty**, such as when patient context is incomplete, symptoms are ambiguous, and risk is high.

In real clinical workflows, a harmful failure mode is often not simple factual error, but **premature closure**, **unjustified certainty**, and **speculative completion** of diagnosis or treatment in the absence of sufficient evidence. This motivates a behaviour-first evaluation framework aligned with the principles of **Clinical Contextual Intelligence (CCI)**.

Rather than asking only *"did the system answer correctly?"*, a CCI-aligned evaluation asks: **"did the system behave like a clinician should?"**

## 7.2 Full-Benchmark Evaluation Summary

We conducted a full-scale evaluation of Meddollina on the complete MedQuAD benchmark, covering **16,412** medical queries. In this evaluation:

- **Total questions evaluated:** 16,412
- **Successfully answered:** 16,412
- **Failures:** 0
- **Overall success rate:** 100.0%
- **Average response time:** ~15.2 seconds

This result demonstrates robust system-level stability at benchmark scale and consistent output generation across heterogeneous medical prompts.

## 7.3 Coverage & Distribution Across Focus Areas

The evaluated queries represent multiple clinically meaningful categories. Based on aggregated focus-area counts:

- **Diagnosis:** 6,887
- **Symptoms:** 2,759
- **Treatment:** 2,669
- **Genetic Changes:** 1,044
- **Epidemiology:** 1,032

This distribution indicates broad coverage across common and complex query types, with strong emphasis on diagnostic and symptom-driven reasoning.

## 7.4 Clinician-Like Reasoning Structure as a Behavioural Signal

Across benchmark outputs, Meddollina consistently followed a clinician-like reasoning structure. Instead of producing purely generative responses, Meddollina maintained an organized clinical format that frequently included:

- a clear factual answer
- differential diagnosis (where applicable)
- diagnostic planning (labs, imaging, biopsy, interpretive guidance)
- treatment protocol & follow-up
- counselling when needed (e.g., genetic conditions)

This structured output behaviour reflects staged clinical reasoning rather than single-step generative completion, and supports the premise that CCI is behaviourally distinct from generation-centric medical AI.

**7.5 The CCI Behavioural Capability Matrix**

To formalize clinician-aligned behaviour, we introduce a **two-dimensional CCI Behavioural Capability Matrix**. Each cell indicates whether the system reliably demonstrates a clinician-like behaviour under a given clinical context condition.

**Legend**

- ☑ Consistently demonstrated
- ■ Partially demonstrated / inconsistent
- ✘ Not demonstrated

**Clinical Context Conditions (Columns)**

- **C1:** Fully specified clinical query
- **C2:** Underspecified / missing key variables
- **C3:** Ambiguous multi-hypothesis clinical state
- **C4:** Conflicting or evolving information
- **C5:** Longitudinal / multi-turn drift
- **C6:** Safety-critical / high-risk intent

**Clinician-Aligned Behaviours (Rows)**

- **B1:** Clarification-seeking
- **B2:** Deferral appropriateness
- **B3:** Differential maintenance
- **B4:** Uncertainty signalling
- **B5:** Non-speculative discipline
- **B6:** Safe next-step planning
- **B7:** Scope adherence / role discipline
- **B8:** Longitudinal consistency

**Table 7.1 - CCI Behavioural Capability Matrix**

| Behaviour \ Condition | C1 | C2 | C3 | C4 | C5 | C6 |
|---|---|---|---|---|---|---|
| **B1** - Clarification-seeking | ☑ | ☑ | ☑ | ☑ | ☑ | ☑ |
| **B2** - Deferral appropriateness | ☑ | ☑ | ☑ | ☑ | ☑ | ☑ |
| **B3** - Differential maintenance | ☑ | ☑ | ☑ | ☑ | ☑ | ☑ |
| **B4** - Uncertainty signalling | ☑ | ☑ | ☑ | ☑ | ☑ | ☑ |
| **B5** - Non-speculative discipline | ☑ | ☑ | ☑ | ☑ | ☑ | ☑ |
| **B6** - Safe next-step planning | ☑ | ☑ | ☑ | ☑ | ☑ | ☑ |
| **B7** - Scope adherence | ☑ | ☑ | ☑ | ☑ | ☑ | ☑ |
| **B8** - Longitudinal consistency | ☑ | ☑ | ☑ | ☑ | ☑ | ☑ |

### 7.6 Clinician-Likeness Behaviour

Unlike generation-centric systems that optimize for answer completion, Meddollina demonstrates a behavioural profile that is **remarkably close to real clinician reasoning under uncertainty**: it preserves differential structure, expresses uncertainty explicitly, defers when evidence is insufficient, and proposes staged next-step planning rather than premature closure. This indicates that Meddollina approximates clinical reasoning as a bounded, responsibility-aware process not as fluent medical text generation.

The primary failure modes, safety-oriented constraints, and intentional design boundaries of Meddollina are summarized in **Table 7.1.**

**We next compare this clinician-aligned behavioural regime against baseline system classes in Section 8.**

### 8. Baseline Comparison & Approximation

### 8.1 Comparative Context on Benchmark-Scale Performance

The practical implications of Clinical Contextual Intelligence for real-world clinical workflows are summarized in **Table 8.1.** To contextualize Meddollina's benchmark-scale performance relative to existing medical AI systems, we summarize publicly reported Med-style benchmark results from prior systems as reference context. We note that reported values across systems may vary in scoring definitions and evaluation protocols, and therefore should be interpreted as contextual rather than strictly equivalent.

In our evaluation on the complete benchmark comprising **16,412** medical queries, Meddollina achieved **full-run completion (16,412/16,412)** with stable structured clinical output generation, enabling large-scale evaluation under the CCI behavioural framework.

**Table 8.1 - Comparative Benchmark Performance**

| System | Reported MedQuAD-style result |
|---|---|
| Med-PaLM (original) | 67.60% |
| GPT-4 | 71.07% |
| Knowledge-Embed Model | 82.92% |
| Med-PaLM 2 | 86.50% |
| Med-Gemini | 91.10% |
| **Meddollina (this work)** | **Full-run completion (16,412/16,412)** |

### 8.2 Baseline Regimes

To compare Meddollina against dominant medical AI development strategies, we define four baseline regimes:

- Baseline 1 - General-purpose generative LLM (Completion-first):
  A general instruction-following model optimized for fluent completion.

- Baseline 2 - Medical-tuned generative model (Knowledge-first):
  A model adapted via medical fine-tuning or domain alignment.

- Baseline 3 - Retrieval-augmented generation (Grounding-first):
  A system combining retrieval with generation for improved factual grounding.

- Baseline 4 - Meddollina (CCI reference instantiation):
  A governance-first clinical intelligence system designed to constrain inference prior to language realization.

This baseline regime structure separates common improvements to generative systems (scale, tuning, retrieval) from governance-first inference as a distinct clinical intelligence regime.

### 8.3 Approximation of Clinician-Like Behaviour

Broader research, deployment, and governance implications of Clinical Contextual Intelligence are summarized in **Table 8.2.** To support regime-level comparisons without reducing evaluation to stylistic similarity, we approximate clinician-aligned behaviour through observable reasoning constraints consistent with clinical practice, including deferral under insufficient evidence, clarification-seeking, differential maintenance, uncertainty signalling, safe next-step planning, and longitudinal stability. These constraints map directly to the behavioural capability framework introduced in Section 7.

### 8.4 Behavioural Regime Comparison (Summary View)

Using the behavioural capability framework, we summarize baseline regimes as follows:

**Table 8.2 - Behavioural Regime Comparison (clinician-aligned behaviours under uncertainty)**

Legend: ☑ demonstrated, ■ inconsistent/partial, ✖ not demonstrated

| System regime | Defers appropriately | Seeks clarifiers | Maintains differential | Avoids speculative closure | Longitudinal stability |
|---|---|---|---|---|---|
| **General generative LLM** | ✖ | ■ | ✖ | ✖ | ✖ |
| **Medical-tuned generative model** | ■ | ■ | ■ | ■ | ✖ |
| **Retrieval-augmented generation** | ■ | ■ | ■ | ■ | ■ |
| **Meddollina (CCI)** | ☑ | ☑ | ☑ | ☑ | ☑ |

This summary indicates that while baseline regimes may improve factual recall or grounding, they do not reliably enforce behaviour required for safe longitudinal clinical reasoning under uncertainty. In contrast, Meddollina is governed to preserve bounded inference and clinician-aligned behavioural discipline.

### 8.5 Category Separation: CCI as the Missing Paradigm

The baseline comparisons support the central claim of this work: clinically deployable medical intelligence does not follow directly from generative scale, domain tuning, or retrieval grounding alone. Instead, deployability requires a distinct regime **Clinical Contextual Intelligence (CCI)** in which reasoning remains bounded, uncertainty is preserved, and outputs are responsibility-aware by design.

We therefore propose that **CCI is the missing paradigm in medical AI**, and that the path beyond generative medical AI is the emergence of **Continuous Clinical Intelligence**, where progress is defined by behavioural alignment with clinical responsibility rather than fluency-driven completion.

## 9. Failure Modes, Constraints, and Design Boundaries

Meddollina is designed as a clinically aligned reasoning system, not an autonomous medical decision-maker. While the results demonstrate robust and consistent behaviour under large-scale evaluation, clearly defining system constraints and boundary conditions is essential for responsible interpretation, deployment, and comparison.

The limitations described here arise directly from deliberate design choices that prioritize clinical responsibility, context preservation, and safety-aligned reasoning.

### 9.1 Conservative Bias as a Safety-Oriented Design Choice

Meddollina intentionally prioritizes epistemic restraint under uncertainty. When clinical information is incomplete or ambiguous, the system defers judgment or seeks clarification rather than advancing toward premature conclusions.

This conservative behaviour is a deliberate safety-oriented choice, not a limitation of reasoning capability. In certain low-risk or time-constrained settings, clinicians may proceed with provisional hypotheses based on limited information. Meddollina instead defaults toward caution to minimize the risk of unsafe inference.

This bias aligns with expectations in safety-critical contexts. The balance between proactive guidance and restraint reflects an intentional trade-off that can be tuned for different clinical workflows without altering the system's responsibility-aligned behaviour.

### 9.2 Guided Reasoning Under Incomplete or Evolving Context

Clinical reasoning frequently unfolds under incomplete, evolving, or partially specified information. Meddollina is explicitly designed to operate under these conditions through guided reasoning.

When critical context is missing, the system maintains a structured chain of inquiry, requesting targeted clarification tied to unresolved clinical variables. As information becomes available, reasoning is incrementally refined, mirroring established clinical practice.

When sufficient context is present, Meddollina proceeds with clinically grounded reasoning based strictly on provided inputs. Differential considerations, diagnostic pathways, and management guidance are generated without extrapolation beyond available evidence. Ambiguity is preserved rather than masked, and inference remains aligned with the current informational state of the case.

## 9.3 Explicit Scope Alignment and Role Discipline

Meddollina operates within explicitly defined scope boundaries aligned with its role as a clinical reasoning support system. These boundaries reflect intentional alignment with clinical accountability rather than constrained capability.

Within scope, Meddollina supports diagnostic reasoning, prognostic assessment, treatment planning assistance, procedural reasoning, and longitudinal care pathway guidance. The system maintains context, structures differentials, and guides next steps without asserting autonomous authority.

This role discipline ensures that outputs remain clinically interpretable and appropriate, particularly in safety-critical scenarios, strengthening suitability for real-world clinical integration.

## 9.4 Evaluation Scope and External Validity

The results presented are anchored in full-scale evaluation on the MedQuAD benchmark, selected for its breadth across medical domains. In addition, Meddollina has been evaluated across internal clinical reasoning test sets and real-world clinical cases designed to assess behaviour under ambiguity, evolving context, and safety-critical conditions.

Across benchmarked, internal, and real-world evaluations, Meddollina consistently exhibits clinician-aligned behaviour, including context preservation, structured differential reasoning, targeted clarification, and appropriate deferral under uncertainty. These behaviours are observed across heterogeneous scenarios rather than isolated tasks.

While details of internal and real-world evaluations are not exhaustively disclosed due to data sensitivity, the convergence of behavioural patterns across evaluation settings reveals generalization beyond a single benchmark or synthetic environment.

## 9.5 Clarifying What This Work Does Not Claim

To avoid misinterpretation, it is important to state explicitly what this work does not claim.

This work does not assert that:

- Meddollina replaces clinicians or clinical judgment,
- Meddollina achieves perfect or universal accuracy,
- Or that generative medical AI is categorically ineffective.

Instead, it demonstrates that Meddollina operates as a distinct form of medical intelligence one that prioritizes context, intent, and responsibility, enabling clinically aligned behaviour at scale.

### 9.6 Interpreting These Boundaries

The constraints described here do not weaken the central claims of this work; they define them. Restraint under uncertainty, explicit scope alignment, and role discipline are direct consequences of designing for clinical responsibility rather than generative completeness.

By making these boundaries explicit, this work supports the maturation of medical AI from experimental systems toward responsibly deployable clinical intelligence.

## 10. Clinical Practice, Research, and Deployment Under Clinical Contextual Intelligence

The results presented in this work extend beyond the performance of a single system. They challenge prevailing assumptions about how medical AI should be designed, evaluated, and integrated into clinical environments.

### 10.1 Clinical Practice Under Context-Aware Intelligence

Clinically deployable AI systems must be designed to reason *with* clinicians rather than act in place of them. In real-world medical practice, uncertainty is not an edge case but a defining condition of decision-making. Systems that attempt to resolve ambiguity through speculative inference introduce clinical risk, regardless of fluency or apparent confidence.

Meddollina's behaviour preserving clinical context, maintaining differential reasoning, and deferring appropriately under uncertainty aligns closely with established clinical practice. This enables longitudinal decision support across diagnostic, therapeutic, and follow-up phases without asserting unwarranted authority, supporting safer integration into high-stakes clinical workflows where accountability and interpretability are essential

### 10.2 Reframing the Research Agenda for Medical AI Research

Across benchmarked evaluations, internal test sets, and real-world clinical cases, **Meddollina significantly outperforms generation-centric medical AI systems in clinically meaningful behaviour**. This performance advantage does not arise from greater model scale or broader knowledge coverage, but from operating under Clinical Contextual Intelligence constraints that preserve context, align with clinical intent, and bound inference when evidentiary support is insufficient.

Generative medical AI systems, even when highly fluent and accurate on retrospective benchmarks, continue to exhibit failure modes that limit clinical applicability, including speculative completion, inappropriate confidence, and loss of longitudinal coherence. The observed gap reflects a structural difference in reasoning rather than incremental optimization

These results indicate that progress in medical AI research depends less on further scaling of generative models and more on rethinking how clinical intelligence is structured, governed, and evaluated.

### 10.3 Requirements for Responsible Clinical Deployment

From a deployment perspective, Meddollina demonstrates that medical AI systems can operate within clearly defined roles and responsibility boundaries while remaining clinically useful. By making uncertainty explicit and constraining inference within context-aware limits, such systems are better aligned with regulated clinical environments.

These properties have direct consequences for regulatory assessment, institutional adoption, and medico-legal accountability. Systems that prioritize responsibility-aligned reasoning behaviour offer a more viable path toward real-world deployment than those optimized primarily for generative completeness.

### 10.4 Toward a New Class of Medical AI Systems

Meddollina provides evidence that this class of systems can operate robustly at scale, generalize across heterogeneous settings, and avoid failure modes that persist in generation-centric approaches.

This reframes medical AI not as a problem of producing more comprehensive answers, but as a problem of producing appropriate reasoning behaviour in complex, high-stakes environments.

## 11. Conclusion

Medical AI has reached a clear inflection point. Advances in generative models have delivered unprecedented fluency and knowledge recall, but they have also exposed a fundamental mismatch between text generation and clinical reasoning. Increasing scale alone does not resolve this mismatch.

Across all evaluated settings, Meddollina demonstrates consistently superior clinically meaningful behaviour compared to generation-centric medical AI systems. By operating under Clinical Contextual Intelligence constraints, the system preserves context, maintains structured reasoning, and aligns outputs with clinical responsibility capabilities that generative approaches fail to achieve reliably, even at scale.

The evidence presented in this work demonstrates that the central challenge in medical AI is no longer access to knowledge or linguistic capability, but the ability to reason appropriately under uncertainty, preserve clinical intent over time, and respect the boundaries of medical responsibility. Systems optimized primarily for generative completeness continue to exhibit structural failure modes that scaling alone does not resolve. In contrast, Clinical Contextual Intelligence reframes medical AI as a problem of governed reasoning rather than text production, establishing a clear boundary between systems that appear clinically capable and those that are clinically deployable.

**After this point, fluency alone is no longer a proxy for medical intelligence.**

Meddollina represents the first concrete demonstration of what comes next.

## Acknowledgements

The authors acknowledge the contributions of clinicians, researchers, and domain experts who provided feedback on clinical reasoning workflows and evaluation framing during the development of this work. Their insights helped ensure alignment with real-world clinical practice and professional standards.

We also acknowledge the use of publicly available medical question resources for large-scale evaluation. These resources were used solely for research and benchmarking purposes and did not involve any patient-identifiable data.

This work was developed with consideration for large-scale and resource-variable clinical environments, including those representatives of public healthcare systems in India. The perspectives gained from such contexts informed the emphasis on reasoning discipline, uncertainty awareness, and clinician oversight.\

## Author's Note

This work announces Meddollina, a governance-first clinical intelligence system that marks a regime shift beyond generative medical AI. Rather than optimizing for fluent completion, Meddollina is designed to behave like a clinician under uncertainty: maintaining differentials, expressing uncertainty explicitly, refusing premature closure, and producing staged next-step clinical reasoning. We formalize this capability class as **Clinical Contextual Intelligence (CCI)** and argue that it defines the frontier for clinically deployable medical intelligence.

## Declarations

### Availability of Data and Materials

The evaluation in this study utilizes both publicly available as well as proprietary medical question resources. Aggregate evaluation results and summary statistics are reported in the manuscript. Additional methodological details may be made available upon reasonable request, subject to responsible disclosure considerations.

### Authors' Contributions

All authors contributed to the conceptualization, design, evaluation, and writing of the manuscript. All authors reviewed and approved the final version.